%% file: main.tex
\documentclass[10pt,twocolumn,letterpaper]{article}

\usepackage{cvpr}              % To produce the CAMERA-READY version

\usepackage[pagebackref,breaklinks,colorlinks,citecolor=cvprblue]{hyperref}
\usepackage[capitalize]{cleveref}

\usepackage[dvipsnames]{xcolor}

\usepackage{algorithm}
\usepackage{algpseudocode}

\usepackage{graphicx}
\usepackage{booktabs}
\usepackage{multirow}
\usepackage{multicol}
\usepackage{colortbl}
\usepackage{scontents}
\usepackage{bbold}
\usepackage{tcolorbox}

\definecolor{cvprblue}{rgb}{0.21,0.49,0.74}

\def\confName{3DV\xspace}
\def\confYear{2025\xspace}

\title{SparseGS: Sparse View Synthesis using 3D Gaussian Splatting}

\newcommand\blfootnote[1]{%
  \begingroup
  \renewcommand\thefootnote{}\footnote{#1}%
  \addtocounter{footnote}{-1}%
  \endgroup
}

\author{
    Haolin Xiong$^{1,3,\dagger,\ast}$ \and
    Sairisheek Muttukuru$^{ 1,\ast}$ \and
    Hanyuan Xiao$^{2,3}$ \and
    Rishi Upadhyay$^{1}$ \and
    Pradyumna Chari$^{1}$ \and
    Yajie Zhao$^{2,3}$ \and
    Achuta Kadambi$^{1}$ \\
    \and
    $^{1}$University of California, Los Angeles
    \and
    $^{2}$University of Southern California 
    \and
    $^{3}$USC Institute for Creative Technologies
}

\begin{document}

\twocolumn[{
\renewcommand\twocolumn[1][]{#1}
\maketitle

\begin{center}
    \vspace{-6mm}
    \setlength{\abovecaptionskip}{4pt}
    \centering
    \includegraphics[width=1.0\linewidth]{figures/teaser.jpg}
    \captionof{figure}{\textbf{The quality of 3DGS~\cite{kerbl3Dgaussians} degrades as the number of input views decreases}, particularly in unbounded scenes. SparseGS significantly improves novel view synthesis quality in sparse-input settings while maintaining fast training and real-time rendering.}
    \label{fig:teaser} 
\end{center}
}
]

% \blfootnote{$\dagger$ Equal contribution.}
% \blfootnote{$\ast$ Corresponding author. \tt\small\{xiongh@ucla.edu\}}
\maketitle
\input{sec/0_abstract}   
\blfootnote{$\ast$ Equal contribution.}
\blfootnote{$\dagger$ Corresponding author. \tt\small\{xiongh@ucla.edu\}}
\input{sec/1_intro}
\begin{figure*}[!h]
    \centering
    \includegraphics[width=\textwidth]{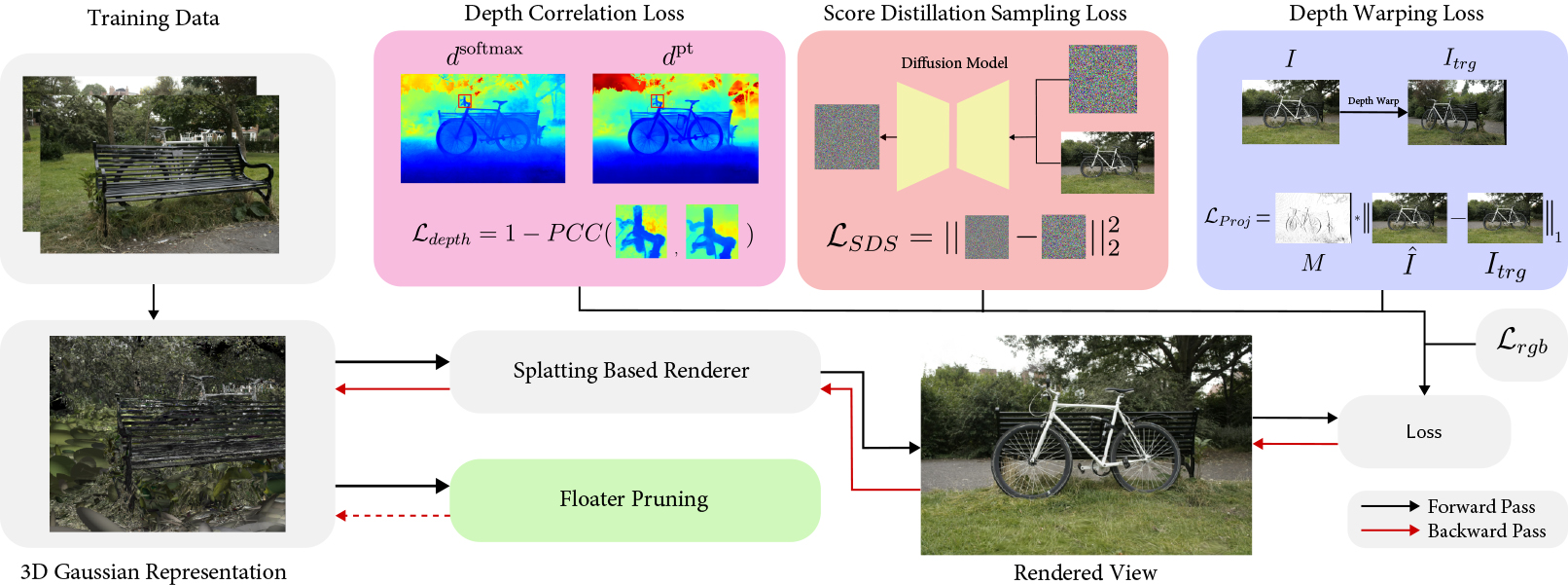}
    \caption{\textbf{Our proposed pipeline incorporates depth priors, diffusion constraints, and a floater pruning technique to improve few-shot novel view synthesis performance.} During training, we render the softmax depth and use Pearson correlation to encourage it to align with $d^{\text{pt}}$ (\cref{sec:depth_corr}). We also generate novel views using the procedure (\cref{sec:diffusion_loss}) and incorporate a Score Distillation Sampling loss. At pre-set intervals, we prune floaters according to the Advanced Floater Removal procedure described in \cref{sec:floater_removal}. % New components we propose are colored while the base 3DGS pipeline is in grey.
    }
    \vspace{-0.5cm}
    \label{fig:model_fig}
\end{figure*}
\input{sec/2_related_work}

\input{sec/3_methods}

\input{sec/4_experiments}
\input{sec/5_conclusion}

{
    \small
    \bibliographystyle{ieeenat_fullname}
    \bibliography{main}
}
\input{sec/X_supp_new}

\end{document}

%% file: sec/0_abstract.tex
\begin{abstract}
3D Gaussian Splatting (3DGS) has recently enabled real-time rendering of unbounded 3D scenes for novel view synthesis. However, this technique requires dense training views to accurately reconstruct 3D geometry. A limited number of input views will significantly degrade reconstruction quality, resulting in artifacts such as ``floaters'' and ``background collapse'' at unseen viewpoints.  In this work, we introduce SparseGS, an efficient training pipeline designed to address the limitations of 3DGS in scenarios with sparse training views. SparseGS incorporates depth priors, novel depth rendering techniques, and a pruning heuristic to mitigate floater artifacts, alongside an Unseen Viewpoint Regularization module to alleviate background collapses. Our extensive evaluations on the Mip-NeRF360, LLFF, and DTU datasets demonstrate that SparseGS achieves high-quality reconstruction in both unbounded and forward-facing scenarios, with as few as 12 and 3 input images, respectively, while maintaining fast training and real-time rendering capabilities. 
 
\end{abstract}

%% file: sec/1_intro.tex
\section{Introduction}
\label{sec:intro}

The challenge of learning 3D representations from 2D images has been a longstanding area of interest, but achieving a balance between efficiency and fidelity remains a persistent challenge. While Neural Radiance Fields (NeRFs) excel in high-quality rendering and effectively represent anisotropic effects in view interpolation, they suffer from long training times, and blurriness if only sparse views are provided as input. The recent development of 3D Gaussian Splatting (3DGS) has substantially reduced the training cost by introducing a more compact, explicit 3D representation coupled with a real-time rendering pipeline. 
% However, because there exist multiple possible 3D representations that can correctly represent the 2D training views, Gaussians may be incorrectly placed, leading to poor rendering quality from viewpoints that differ from the training views, especially when the number of training views is limited.
However, 3DGS still suffers from artifacts caused by the inherent ambiguity in projection from 3D to 2D posed by sparse input views.

Artifacts in 3DGS, such as ``floaters'' (high-density floating regions due to misplaced Gaussians) and ``background collapse'' (caused by Gaussians being misplaced at incorrect depths, resulting in background Gaussians appearing in the foreground) tend to be more pronounced than those in NeRFs. These issues are further exacerbated when the training set lacks substantial scene coverage, such as in multi-view unbounded scenes~\cite{barron2022mipnerf360} (referred as 360-degree scenes in the rest of this paper). Extensions of 3DGS, such as ~\cite{zhu2023FSGS,li2024dngaussian}, have attempted to incorporate depth priors as geometry supervisions or regularizers, but they still fail to resolve the problem of floaters, particularly in unbounded scenes. We observe that naively applying the same alpha-blending equation for rendering depth can cause gradients to propagate to the wrong Gaussians, adversely affecting quality. Additionally, only providing extra guidance from the training views does not mitigate the problem of overfitting, thus failing to address background collapse issues in sparse-input settings.

Our objective is to accurately reconstruct 360-degree unbounded 3D scenes using as few as 12 input images. To address the aforementioned ``floater'' and ``background collapse'' issues in 3DGS representations, we introduce three key modules. First, we propose two novel depth rendering techniques (softmax-scaling depth and mode-selection depth) that go beyond the widely used alpha-blending depth to more effectively manage floaters. Next, we introduce a module designed to tackle background collapse by leveraging a 2D generative diffusion prior~\cite{poole2022dreamfusion, liu2023zero1to3} and depth warping~\cite{4547850, Xu_2022_SinNeRF}. Finally, capitalizing on the explicit nature of 3DGS representations, which allows for direct scene manipulation, we present a floater pruning procedure to identify and eliminate undesired Gaussians. Combined, our pipeline achieves state-of-the-art (SOTA) performance in sparse-input novel view synthesis (NVS) problems, not only on forward-facing datasets but also on 360-degree unbounded scenes, a scenario that most current few-shot techniques~\cite{wang2023sparsenerf, Niemeyer2021Regnerf, li2024dngaussian} struggle to handle effectively.\\
\vspace{2pt}\noindent\textbf{In summary, our contributions are:}
\begin{enumerate}
    \item We propose a novel framework, \textbf{SparseGS}, for training coherent and robust 3D Gaussian representations from limited inputs, outperforming SOTA methods in sparse view synthesis.
    \item SparseGS addresses the ``floater'' issue by introducing mode-selection and softmax-scaling depth rendering techniques.
    \item SparseGS introduces an Unseen Viewpoint Regularization module, which mitigates overfitting by regularizing 3DGS training at viewpoints different from the input views. Empirically, the module reduces ``background collapse'' in sparse-input settings.
    \item SparseGS also introduces an explicit and adaptive operator to further prune undesirable floating artifacts in 3DGS.
\end{enumerate}

%% file: sec/2_related_work.tex
\section{Related Work}
\label{sec:related_work}
We focus on neural representations of 3D scenes, particularly radiance fields, considering their offered detail level and scalability in novel view synthesis Additionally, we explore related works aimed at enhancing rendering quality and addressing artifacts when only sparse view images are provided as input.

\vspace{2pt}\noindent\textbf{Radiance Fields.} Neural radiance field (NeRFs) as a 3D scene representation was first introduced by~\cite{mildenhall2020nerf}. NeRF learns a continuous field of density and color that naturally interpolates high-dimensional appearance features in an differentiable way. The volumetric rendering equation is 
\begin{align}
C = \sum_{i=1}^{N} T_i \sigma_i c_i,
\end{align}
\vspace{-0.5cm}
with 
\vspace{-0.5cm}
\begin{align}
    \alpha_i=(1-\exp(-\sigma_i \delta_i)), T_i=\prod_{j=1}^{i-1}(1-\alpha_i),
\end{align}
where $C$ is the aggregated color along a ray, $T_i$ is the transmittance of a sampled point $i$ along the ray, $\sigma_i$ is the density of $i$, and $c_i$ is the color of $i$. A flurry group of extension works introduce more challenging datasets~\cite{mildenhall2022rawnerf, levis2022gravitationally}, improve training speed~\cite{mueller2022instant} or quality~\cite{barron2021mipnerf}, and apply NeRF to other tasks~\cite{zhang2021nerfactor, weng_humannerf_2022_cvpr}. Among them, Mip-NeRF~\cite{barron2021mipnerf} introduced the concept of anti-aliasing to NeRFs by rendering conical frustums rather than rays. As a result, Mip-NeRF is able to anti-alias renderings at consistently high quality at focal lengths different from training views. Mip-NeRF 360~\cite{barron2022mipnerf360} followed up Mip-NeRF that specifically tackled the bottleneck in 360\textdegree~unbounded scene representation. In this work, we evaluate on both forward-facing scenes and 360\textdegree~unbounded scenes in challenging sparse-view input task.

% While NeRFs rely on a neural network to represent the radiance field, 3D Gaussian Splatting~\cite{kerbl3DGaussians}, is a recent technique for view synthesis that replaces the neural network with explicit 3D Gaussians. These Gaussians are parameterized by their position, rotation, scaling, opacity, and a set of spherical harmonics coefficients for view-dependent color. They are then rendered by projecting the Gaussians down to the image plane using splatting-based rendering techniques~\cite{964490, zwicker2001surface} and then alpha blending the resulting colors and opacities. Models trained using the same reconstruction loss as NeRFs. Replacing ray-tracing and neural network process of NeRF-based methods with splatting and direct representation process offers significant run-time improvements and allows for real-time rendering during inference. In addition, 3D Gaussian representations are explicit rather than the implicit representations of NeRFs, which allows for more direct editing and easier interpretability. We leverage this property for our technique which identifies and directly deletes floaters.

3D Gaussian Splatting (3DGS)~\cite{kerbl3Dgaussians} introduced a point-based differentiable 3D representation that achieves competitive rendering quality, training time and also real-time rendering in high resolution. The points with volume (or splats) are parameterized by their position, rotation, scaling, opacity, and a set of spherical harmonics coefficients for view-dependent color. Specifically, the geometry of splats is defined by a full 3D covariance matrix that is composed by a scaling matrix and a rotation matrix. The differentiable settings enable training with simple color supervision same as NeRFs. 3DGS advances NeRF in scalability of the representation, where NeRF cannot encode large scenes with limited number of parameters in a MLP. In addition, 3D Gaussian representations are explicit rather than the implicit representations of NeRFs, which allows for more direct editing and easier interpretability. We leverage this property for our technique which identifies and directly deletes floaters.

Recent works have built on top of 3D Gaussian Splatting to perform a variety of downstream tasks including text-to-3D generation~\cite{tang2023dreamGaussian, yi2023Gaussiandreamer}, dynamic scene representation~\cite{luiten2023dynamic, wu20234d}, and animating humans~\cite{zielonka2023drivable}. 

\vspace{2pt}\noindent\textbf{Few Shot Novel View Synthesis.} The problem of novel view synthesis from few images has received significant interest as many view synthesis techniques can require a prohibitively high number of views for real-world usage. Early techniques~\cite{tucker2020single, zhou2018stereo, srinivasan19} leverage multi-plane images (MPIs), which represent images by sub-images at different depths, to re-render depth and color from novel view points using traditional transformations. More recent techniques are built on top of the NeRF framework and tend to tackle the problem in one of two ways: The first set of methods introduce constraints on the variation between views. An early example of this type of method was DietNeRF~\cite{Jain_2021_ICCV} which added constraints to ensure that high-level semantic features remained the same from different views since they contained the same object. Another example is RegNeRF~\cite{Niemeyer2021Regnerf} which applies both color and depth consistency losses to the outputs at novel views. The second set of methods approach the problem by adding depth priors to novel views to regularize outputs. SparseNeRF~\cite{wang2023sparsenerf} falls into this category and uses a pre-trained depth estimation model to get pseudo-ground truth depth maps which are then used for a local depth ranking loss. They additionally apply a depth smoothness loss to encourage rendered depth maps to be piecewise-smooth. DSNeRF~\cite{kangle2021dsnerf} also uses additional depth supervision, but uses the outputs from a Structure-From-Motion(SFM) pipeline (typically COLMAP~\cite{schoenberger2016sfm}) instead of a pre-trained model. Other than the techniques above, Neo 360~\cite{irshad2023neo} uses tri-planes to represent a bounded scene more efficiently. \cite{li2024dngaussian} tackles the sparse-view 3DGS reconstruction with depth regularization and monocular depth estimation. Concurrent work~\cite{zhu2023FSGS} introduces a more advanced densification strategy tailored to 3DGS and leverages a similar zero-shot depth estimator~\cite{Ranftl2020,Ranftl2021}. To the best of our knowledge, \cite{zhu2023FSGS, irshad2023neo} are the only existing methods that explicitly tackle the problem of 360\textdegree~few-shot novel-view synthesis.

%% file: sec/3_methods.tex
\newcommand{\dalpha}{$d^{\text{alpha}}$}
\newcommand{\dmono}{$d^{\text{pt}}$}
\newcommand{\dsoftmax}{$d^{\text{softmax}}$}

\section{Methods}
\label{sec:methods}
\vspace{2pt}\noindent\textbf{Overview.}
Our method consists of three key components designed to function cohesively to improve view consistency and depth accuracy in novel view synthesis: a depth correlation loss, an Unseen Viewpoint Regularization (UVR) module, and a floater pruning operation. In the following section, we first discuss three different ways of rendering depth from 3DGS scenes, followed by a patch-based depth loss. Then, we dissect the UVR module into two parts: a Score Distillation Sampling (SDS) loss and a depth warping loss, which are designed for regularizing viewpoints distant and close to training cameras, respectively. Finally, we introduce a floater pruning procedure, which utilizes depth to identify and remove misplaced Gaussians (``floaters''). \cref{fig:model_fig} showcases a high-level architecture of our pipeline. 

% \subsection{Differentiable Depth Rendering from 3DGS}
\subsection{Mode-selection \& Softmax-scaling Depth Rendering}
\label{sec:render_depth}
\begin{figure}
  \centering
  \includegraphics[width=0.4\textwidth]{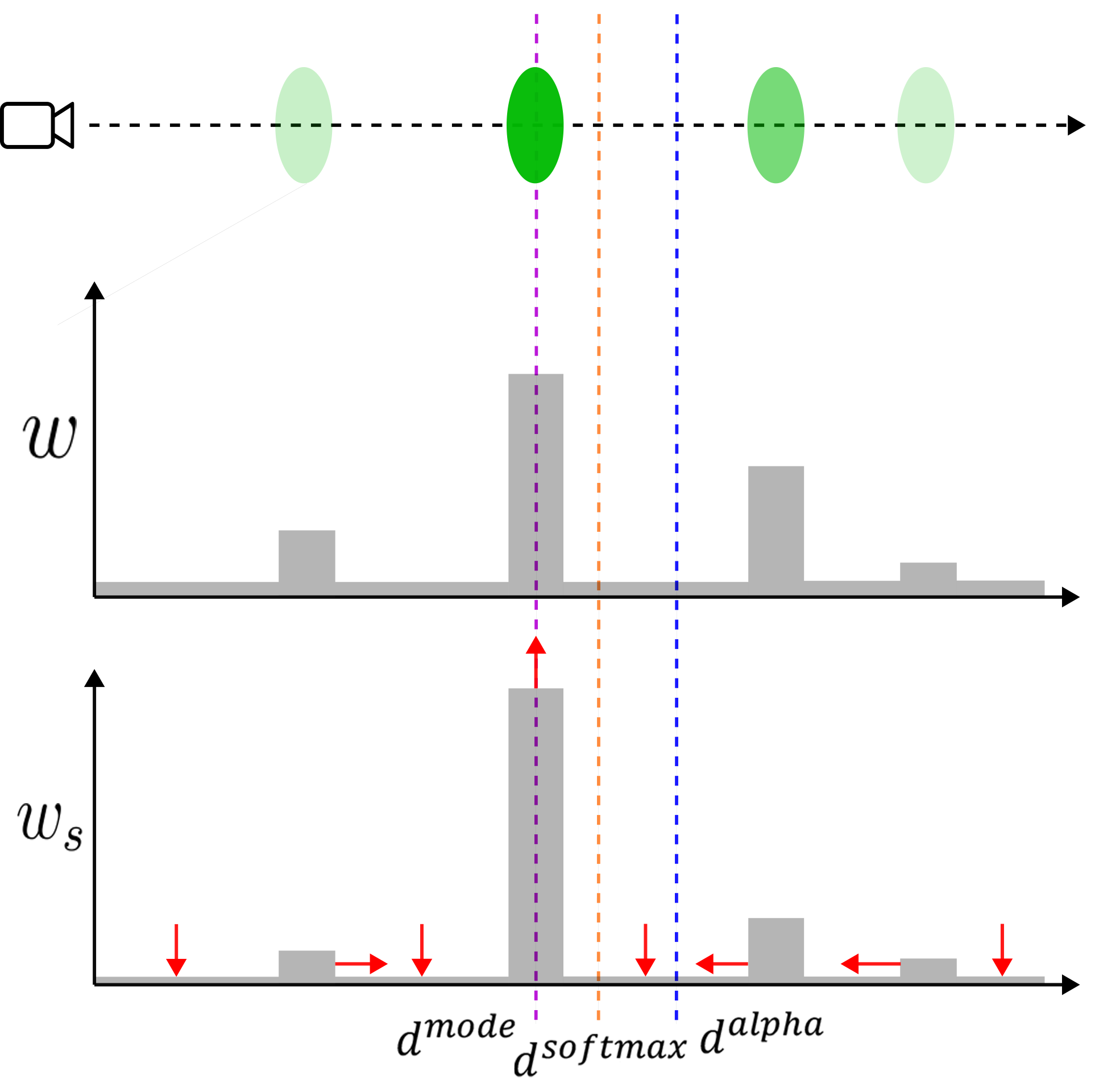}
  \caption{\textbf{A Demonstration of the Three Kinds of Depth.} The weights $w_i$ are shown at the top, with the weights after applying softmax displayed as $w_s$ below. Although a single Gaussian may have the highest weight, other nearby low-weight Gaussians still influence the depth calculation. By applying softmax-scaling, depth accumulation is biased toward Gaussians with higher weights. The Depth Pearson Correlation is less likely to distort the actual depth by manipulating the opacity of floaters. By the end of training, our model consolidates the low-weight Gaussians into a single Gaussian point at the correct depth, ensuring that all depth variations align.}
  \label{fig:logsoftmax_depth}
\end{figure}

% Many of the components we introduce rely on depth maps rendered from the 3D Gaussian representation. To compute these, we use three different techniques, each with unique properties: alpha-blending, mode-selection, and softmax scaling.
Alpha-blending is a widely used technique in NeRFs~\cite{mildenhall2020nerf} for rendering depth maps. Applying the same method to 3DGS, the alpha-blending depth at pixel $(x,y)$, denotes as $d^{\text{alpha}}_{x,y}$, is calculated as:
\begin{align}
d^{\text{alpha}}_{x,y} = \sum_{i=1}^N T_i \alpha_i d_i
\end{align}
where $T_i$ is the accumulated transmittance for the $i$-th Gaussian, $\alpha_i$ is the alpha-compositing weight, and $d_i$ is the depth of the Gaussian. In this rendering approach, the depth value of a pixel is influenced by all Gaussians along their corresponding ray. As a result, the model might adjust the transmittance values of incorrectly placed Gaussians instead of altering their positions, making the 2D depth map appear accurate from the training angles despite the incorrect underlying 3D geometries. So, we introduce two different ways of rendering depth from 3DGS, which are used in later sections.
% \begin{figure}
%     \centering
%     \includegraphics[width=0.49\textwidth]{figures/mode_fig.pdf}
%     \caption{\textbf{An illustration of how $d^{\text{alpha}}$ and $d^{\text{mode}}$ can disagree.} On top, we see an example ray with 4 Gaussians along it. On bottom, we see values of $T_i$, $\alpha_i$, and $w_i$ along this ray. Note the upward and downward axes both represent positive values and are shown this way to reduce clutter. The low opacity Gaussians far out pull the alpha-blended depth back slightly, leading to ambiguity about the true depth. With more Gaussians along a ray, this problem can be exacerbated.}
%     \label{fig:depth_fig}
% \end{figure}

\vspace{2pt}\noindent\textbf{Mode-selection.}
In mode-selection, the depth of the Gaussian with the largest contributing $w_i=T_i\alpha_i$  represents the depth of that pixel. The mode-selected depth of a pixel can be written as:
\begin{align}
d^{\text{mode}}_{x,y} =d_{\arg \max_i \left( w_i \right)}.
\end{align}

\noindent
Intuitively, the mode-selected depth chooses the highest-opacity Gaussian, while the alpha-blended depth takes into account all Gaussians along the imaginary ray from the camera. A crucial insight for building our pruning operator is that $d^{\text{mode}}$ and $d^{\text{alpha}}$ should ideally be the same. Consider the toy setting on the left of~\cref{fig:logsoftmax_depth}. In this scenario, $d^{\text{mode}}$ corresponds to the depth of the second Gaussian from the left, as it has the highest $w_i$. However, $d^{\text{alpha}}$ appears slightly behind this Gaussian due to the influence of low-weight Gaussians with greater depths. This discrepancy can create ambiguity regarding the true depth. We interpret this ambiguity as a measure of the uncertainty of 3DGS at a given point and leverage it to guide our pruning operator introduced later.

\vspace{2pt}\noindent\textbf{Softmax-scaling.}
While rendering depth using the mode identifies the most significant Gaussians contributing to the depth, the $\arg \max$ operator restricts the gradient during backpropagation flow to only the Gaussian with the highest $w_i$. This approach can be problematic when a Gaussian closer to the viewer is translucent, as the weights of farther Gaussians should be non-zero. In order to overcome this limitation, we further introduce a softmax-scaling to modify alpha-blending: 
\begin{align}
d^{\text{softmax}}_{x,y} = \log \left( \frac{ \sum_{i=1}^N w_i e^{\beta w_i}  d_i}{\sum_{i=1}^N w_i e^{\beta w_i} } \right),
\end{align}

\noindent
where the $\beta$ parameter allows us to modulate the softmax temperature and thereby, to choose a desired amplification of highly weighted Gaussians. 
Note that:
\begin{align}
\lim_{\beta \to 0}d^{\text{softmax}}_{x,y} = \log(d^{\text{alpha}}_{x,y}), \\
\lim_{\beta \to \infty}d^{\text{softmax}}_{x,y} = \log(d^{\text{mode}}_{x,y}).
\end{align}

With the softmax-scaling add-on, we can approximate the mode depth while still propagating gradients to Gaussians off the mode.

\subsection{Patch-based Depth Correlation Loss}
\label{sec:depth_corr}
We compute pseudo-ground truth depth maps using pre-trained depth estimation models on the training views. We opt for the Marigold monocular depth estimator~\cite{ke2023repurposing}, though any pre-trained depth estimation model~\cite{Ranftl2020, Ranftl2021, Miangoleh2021Boosting, stan2023ldm3d} could work. We refer to the depth from this pre-trained model as $d^{\text{pt}}$.

Since the monocular estimation model predicts relative depth, while alpha-blending, softmax-scaling, and mode-selection depths are COLMAP-anchored, directly applying an L2 loss, such as mean squared error (MSE), would be ineffective. One option is to estimate scale and shift parameters to align the two depth maps in metric space. However, this transformation is not guaranteed to be constant across all pixels, and a naive alignment could introduce additional unwanted distortion. Instead, we adopt Pearson correlation across image patches to compute a similarity metric between depth maps. This approach is derived from a similar intuition as the depth ranking losses proposed by prior work~\cite{wang2023sparsenerf}, in that they both leverage relative depth to ascertain global depth. However, rather than comparing the ranks of two selected pixels per iteration, we compare between entire patches, allowing the model to influence larger portions of the image and learn more local structures. The Pearson correlation coefficient encourages patches at the same location in both depth maps to have high cross-correlation, regardless of variations in depth value ranges. At each iteration, we randomly sample $N$ non-overlapping patches to compute the depth correlation loss as: 

\begin{gather}
\mathcal{L}_{\text{depth}} = \frac{1}{N} \sum^{N}_i 1 - \text{PCC}(p^{\text{softmax}}_i, p^{d^{\text{pt}}}_i)
% PCC(X,Y) = \frac{\mathbb{E}[XY] - \mathbb{E}[X]\mathbb{E}[Y]}{\sqrt{\mathbb{E}[Y^2] - \mathbb{E}[Y]^2} \sqrt{\mathbb{E}[X^2] - \mathbb{E}[X]^2}}
\end{gather} where $p^{\text{softmax}}_i \in \mathbb{R}^{S^2}$ denotes the $i$-th patch of \dsoftmax and $p^{\text{pt}}_i \in \mathbb{R}^{S^2}$ denotes the $i$-th patch of $d^{pt}$, with the patch size $S$ being a hyper-parameter empirically chosen to be 1/100 of the image resolution. Additionally, a global Pearson depth loss should be applied to prevent depth discontinuities from appearing at patch edges. Intuitively, this loss works to align the softmax-scaling depth maps of the Gaussian representation with $d^{\text{pt}}$ while avoiding the problem of inconsistent scale and shift.

\subsection{Unseen Viewpoints Regularization (UVR)}
% Ambiguities in 3D arise from excessive overfitting to the training views, which is particularly problematic in sparse-input settings. In this section, we introduce a module to regularize 3DGS from unseen viewpoints. The SDS Loss distills knowledge from a large 2D foundation model~\cite{rombach2021highresolution} into viewpoints far from the training cameras, while the Depth Warping Loss aims to regularize views close to the training cameras.
In this section, we propose two regularization methods to improve reconstruction from novel viewpoints:\\
1). Score Distillation Sampling (SDS) Loss: This loss leverages a large vision model~\cite{rombach2021highresolution} to guide training from viewpoints that are distant from the original training cameras, ensuring that the renders appear more natural.\\
2). Depth Warping Loss: This loss utilizes monocular-estimated depth to warp training images to nearby angles, effectively generating additional pseudo-training cameras from the re-projected views.

\subsubsection{Score Distillation Sampling Loss}
\label{sec:diffusion_loss}
\begin{figure}[h!]
    \centering
    \includegraphics[width=0.46\textwidth]{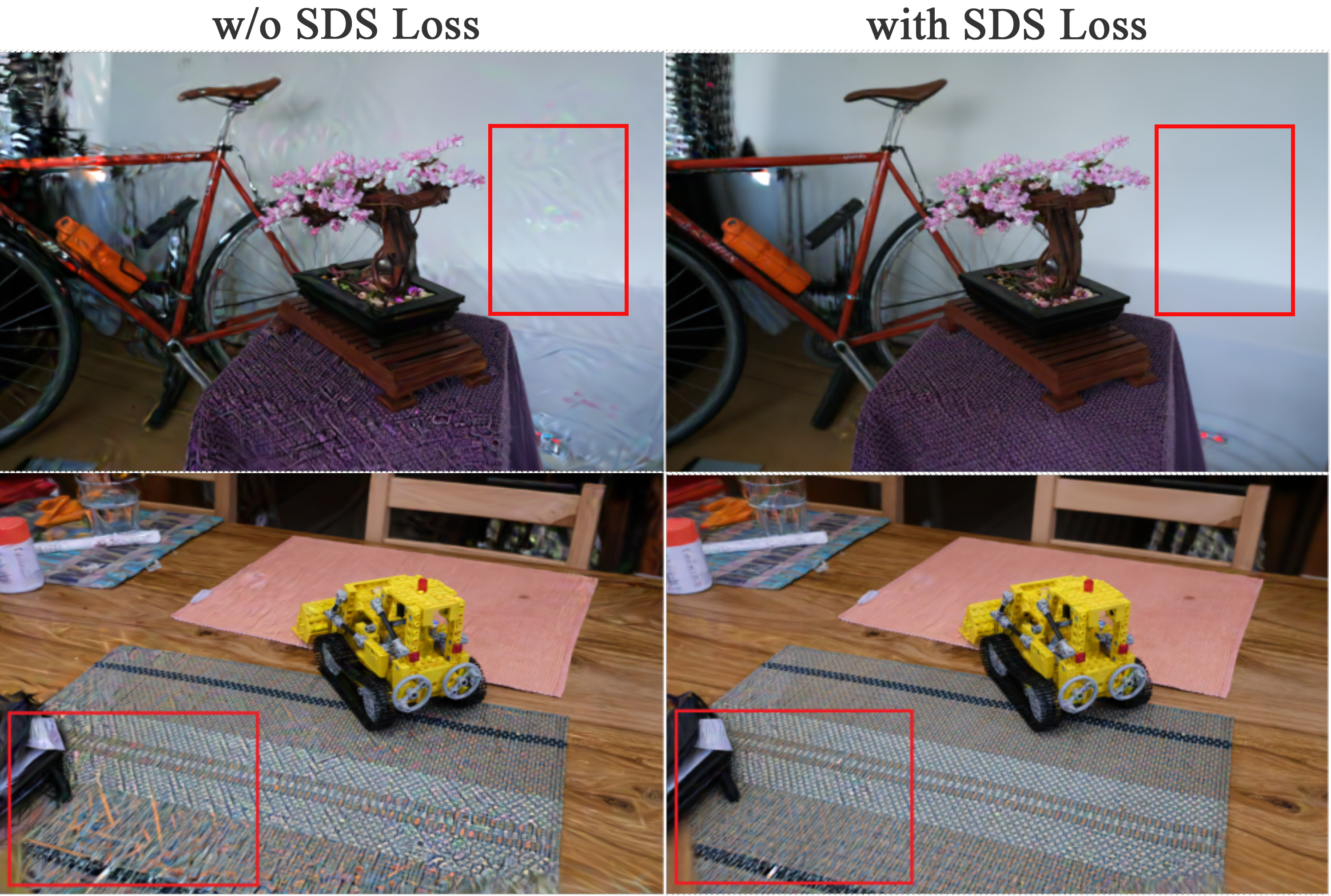}
    \caption{\textbf{Illustration of benefits from the SDS loss.} While the scene structure is well preserved, the high-frequency noise in both geometry and texture is significantly reduced (red box).}
    \label{fig:sds}
\end{figure}

In the sparse-view setting, Gaussians that are well constrained under input viewpoints often appear as small fragmentation rendered from other sampled viewpoints. This issue is caused by misplaced Gaussians or loosely constrained high-dimensional color representations. Simple smoothing techniques, such as Laplacian smoothing, lead to loss of sharpness in detail. Inspired by recent diffusion models~\cite{poole2022dreamfusion, rombach2022high, ho2020denoising, po2023state, jain2021dreamfields, Xu_2022_neuralLift} and Score Distillation Sampling (SDS)~\cite{tang2023dreamGaussian} for zero-shot 3D reconstruction~\cite{liu2023zero1to3, hong2023lrm, liu2023one, pix3d}, we propose using a pre-trained 2D diffusion prior to refine 3DGS reconstruction. The diffusion prior is expected to preserve meaningful image details (in-distribution modeled by diffusion model) while removing non-photorealistic artifacts (out-of-distribution modeled by diffusion model). 

Specifically, we implement a strategy that samples random viewpoints around the center of scene estimated from input cameras. Then, the renderings at the sampled viewpoints are encoded and decoded by the diffusion model, where the predicted noise is then supervised with our SDS loss, formulated as:
\begin{align}
    \hat{I} = \mathcal{N}(\sqrt{\hat{\alpha}}I', (1 - \hat{\alpha})\mathbf{I}), \\
    \mathcal{L}_{\text{SDS}} = \nabla_{G} \mathbb{E}[(\epsilon_{\phi}(\hat{I}_p; \tilde{I_p}) - \epsilon) \frac{\partial \hat{I_p}}{\partial G}],
\end{align}
where $G$ represents the parameters of our Gaussian representation, $\hat{a}$ represents the cumulative product of one minus the variance schedule, $\epsilon$ is the sampled noise, and $\epsilon_{\phi}(\cdot)$ is the predicted noise by the diffusion model. $\hat{I}_p$ denotes the rendered image at camera pose $p$ with added noise, encoded by the diffusion model, and $\tilde{I_p}$ is the denoised image. We compare rendering results with and without our SDS loss in ~\cref{fig:sds}, where the module successfully removes high-frequency artifacts while leaving the scene structure untouched.

% Note that the usage of our SDS loss greatly differs from most prior uses of diffusion models for sparse view synthesis in a way that it does not hallucinate large blank regions. 

% While most prior techniques leverage SDS to fill-in large blank regions or hallucinate new views~\cite{chen2023Gaussianeditor,2408.00083,2305.11588} (e.g. hallucinate back views while only front views are present in the training set), we use the SDS loss to make renders appear more natural, leveraging the fact that the pre-trained diffusion model is trained with real-world images. See~\cref{fig:sds} for examples.

\subsubsection{Depth Warping Loss}
\label{sec:depth_warping}
% Most 3D reconstruction models operate by fitting to the training views, with the expectation that it will learn the underlying geometry of the scene. However, in sparse-view settings, excessive overfitting to limited views ~\cite{Jain_2021_ICCV, Niemeyer2021Regnerf} leads to background collapse even with minimal changes in the viewing angle. This undermines the model's ability to generalize and accurately reconstruct the scene from new perspectives. In our study, we employ image re-projection, utilizing established depth warping techniques~\cite{4547850, Xu_2022_SinNeRF, kangle2021dsnerf} to augment the available training data by re-projecting images to nearby viewpoints.

3DGS operates by fitting to the training views, with the expectation that it will learn the underlying geometry of the scene. However, in sparse-input settings, the model tends to excessively overfit to the limited data, leading to significant background collapse even with slight changes in viewing angles. Regularizing the training views with depth prior helps but is not enough in extreme sparse-view scenarios. To address this, we employ image re-projection, utilizing established depth warping techniques~\cite{4547850, Xu_2022_SinNeRF} to augment the training data by re-projecting images to nearby viewpoints. 

Unlike previous methods~\cite{kangle2021dsnerf} that rely on depth maps generated from COLMAP, which often suffer from noise due to spurious correspondences, noisy camera parameters, or poor COLMAP optimization, we have discovered that scaling the relative range of monocular-estimated depth maps to match the range of rendered alpha-blending depth maps is sufficient for depth warping. Consequently, the quality of the warping heavily depends on the quality of $\min_i d^{\text{alpha}}_i$, $\max_i d^{\text{alpha}}_i$, and $d^{\text{pt}}$. However, in practice, with the current monocular estimation model~\cite{ke2023repurposing}, we have observed that most areas in the warped results are stable, provided that the aforementioned Pearson depth loss has converged to a reasonable level.

Mathematically, we define our image re-projection as follows: For pixel $p_i(x_i,y_i)$ in training image $I_{\text{src}}$, the warping to the corresponding pixel $p_j(x_j,y_j)$ at an unseen viewpoint $I_{\text{trg}}$ can be formulated as:
\begin{align}
    p_j = K_{\text{trg}}T(K^{-1}_{\text{src}}Z_ip_i),
\end{align}
where $Z_i$ represents the monocular-estimated depth map scaled to the range of $\min_i d^{\text{alpha}}_i$ and $\max_i d^{\text{alpha}}_i$. $K_{\text{trg}}$ and $K_{\text{src}}$ are the intrinsic matrices of the source and the target camera, and $T$ refers to the transformation between camera poses from viewpoint $I_{\text{src}}$ to $I_{\text{trg}}$. A warp mask $M$ indicating which pixels are validly warped is generated along the process. Finally, the Depth Warping loss $\mathcal{L}_{\text{Proj}}$ is formulated as:
\begin{align}
    \mathcal{L}_{\text{Proj}} = M * \mathcal{L}_1(\hat{I}, I_{\text{trg}}).
\end{align}

\subsection{Advanced Floater Pruning}
\begin{figure}[tb]
    \centering
    \includegraphics[width=0.48\textwidth]{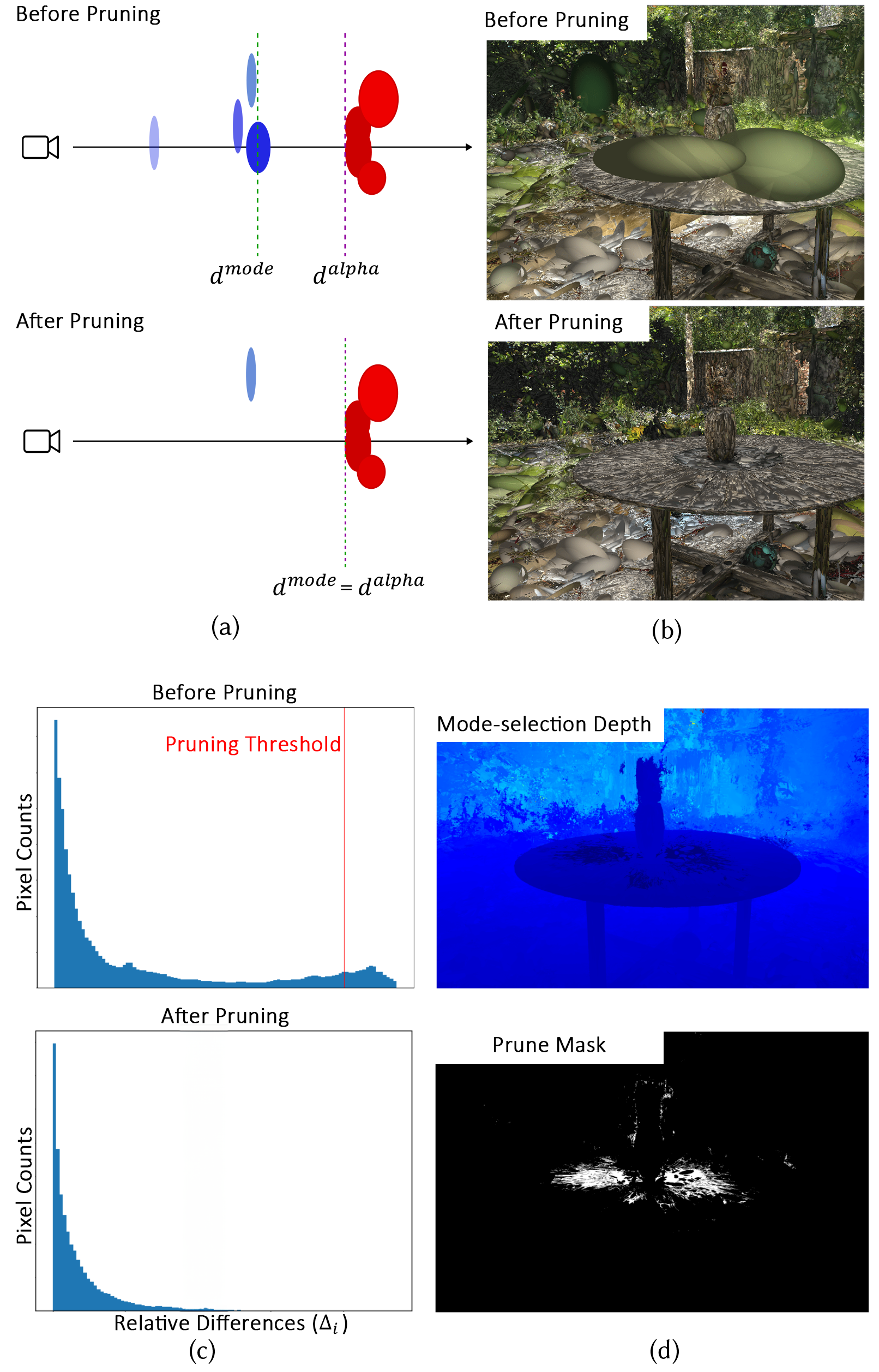}
    \caption{\textbf{The proposed floater pruning technique removes Gaussians at inaccurate depths.} An example (a) demonstrates our pruning method: before pruning, there are floaters (blue) in front of the Gaussians at the object surface (red) and therefore, $d^{\text{mode}}$  $d^{\text{alpha}}$ are not aligned. The pruning method removes all Gaussians on that pixel before the mode and as a result, $d^{\text{mode}} = d^{\text{alpha}}$. Applied to the garden scene (b), the pruning method removes large floaters at the center and left of the scene. We see the relation between the bimodality of the histogram of relative differences between $d^{\text{mode}}$ and $d^{\text{mode}}$, before and after the pruning operation in (c), with the appropriate pruning cutoff points indicated by the red vertical line. (d) The prune mask $F_i$ (bottom) is obtained from thresholding the relative differencing of the mode depth $d^{\text{mode}}$ (top) at the cutoff points.}
    \label{fig:floater_pruning}
    \vspace{-0.5cm}
\end{figure}

\label{sec:floater_removal}
% \vspace{-8pt}
%Although the model is trained with the depth correlation loss, optimizing the softmax depth alone does not solve the problem of ``floaters'' entirely. An example image with floaters is shown in~\cref{fig:floater_pruning}(b). We propose a novel pruning operator which leverages the explicit representation of 3DGS to remove floaters and encourages the model to re-learn those regions of the training views correctly.
%In 3DGS, ``floaters'' are relatively low opacity Gaussians positioned close to the camera center. They do not appear prominently when we render the softmax depth of a scene because they are ``averaged out'', but they are prominent in the mode-selected depth. Even with the depth loss, there are regions where the mode and alpha depth are severely misaligned. This difference can be largely attributed to the fact that 3DGS stops rasterizing after a transmittance threshold has been reached. The floater gaussians close to the camera decay this transmittance budget early and stop the gaussians from converging at the true depth. To increase the rasterization depth in alpha-mode misaligned regions we threshold the areas of misalignment to generate a floater mask $F$ for each input sparse view. % The floater mask is ...

Because the softmax depth loss is a soft constraint, there may exist regions where $d^{\text{mode}}$ and $d^{\text{alpha}}$ do not align. As a result, some floaters may remain along the rays of the input views. Therefore, we propose a novel pruning operator to remove the Gaussians at false modes at the end of training. 

We generate a mask $F_i$ for each training view $i$ by calculating  $\Delta_i$ , which represents the per-pixel relative difference between the depth obtained from mode selection and the depth from alpha blending. Upon examining the distribution of $\Delta_i$, we observed that images containing numerous floaters typically exhibit bimodal histograms. This is because floaters are usually positioned significantly away from the true depth. Conversely, images without floaters tend to have more unimodal histograms. This difference is illustrated in \cref{fig:floater_pruning}(c). Based on this observation, we apply a threshold to  $\Delta_i$  to make the distribution more unimodal, thereby minimizing the presence of floaters. An example of a floater mask is depicted in \cref{fig:floater_pruning}(d). After generating the mask, we proceed to identify and prune all Gaussians, including the mode Gaussian. The impact of this pruning is demonstrated in \cref{fig:floater_pruning}(b). The complete pruning algorithm is detailed in the supplementary materials.

%% file: sec/4_experiments.tex
\section{Experiments}
\label{sec:results}

% \paragraph{Mip-NeRF 360 Dataset}
\label{sec:mipnerf360_res}

\newcommand{\widthFigure}{.19\textwidth}
\newcommand{\widthqualFigure}{\textwidth}
\newcommand{\heightSpace}{1pt}
\newcommand{\compfigure}{0.2\textwidth}

%1red \cellcolor[HTML]{FFCCC9}
%2orange \cellcolor[HTML]{FFE4CF}
%3yellow \cellcolor[HTML]{FFFFD4}

\subsection{Experimental Settings}
\vspace{2pt}\noindent\textbf{Datasets.} We conduct our experiments on three datasets, categorized into two settings: 1) the Mip-NeRF360~\cite{barron2022mipnerf360} dataset, which features seven challenging 360\textdegree~scenes; 2) the LLFF~\cite{mildenhall2019llff} and DTU~\cite{jensen2014large} datasets, which contain forward-facing scenes. For the Mip-NeRF360 dataset, we use every eighth image as testing view and evenly sample 12 or 24 views from the remaining views as the training set. For the LLFF and DTU datasets, we follow the training and evaluation protocols established in RegNeRF~\cite{Niemeyer2021Regnerf}. Following conventions, all input images are downscaled to 1/4 of the original height and width.

% \vspace{2pt}\noindent\textbf{Baselines.} We compare SparseGS with several NeRF-based and 3DGS-based NVS methods on these datasets. RegNeRF~\cite{Niemeyer2021Regnerf}, SparseNeRF~\cite{wang2023sparsenerf}, DNGaussian~\cite{li2024dngaussian}, and FSGS~\cite{zhu2023FSGS} are designed to handle sparse-input settings, while Mip-NeRF360~\cite{barron2022mipnerf360} is a high-performance 360-degree NVS method. We use PSNR, SSIM, and LPIPS scores as quantitative metrics.

\vspace{2pt}\noindent\textbf{Implementation Details.} 
% We implemented SparseGS using PyTorch. 
We obtain the camera calibration of the input views and the initial point cloud for 3DGS using COLMAP~\cite{schoenberger2016sfm,schoenberger2016mvs}. Specifically, the initial point cloud is output from the multi-view stereo (MVS) step. % Camera poses are obtained using COLMAP Structure-From-Motion(SFM)~\cite{schoenberger2016sfm}. All 3DGS-based methods are initialized with COLMAP Multi-View-Stereo(MVS)~\cite{schoenberger2016mvs} point clouds generated from only the training views. We have also tested the performance of SparseGS and FSGS under different initial point clouds. Further analysis are shown in section ~\cref{sec:mipnerf360}. 
%For Depth Warping loss, each training image is warped 2 times in each direction (4 times total) for 1.5 degrees around the average up-vector of the training cameras. 
% The SDS loss, which use Stable-Diffusion-2.1 as backbone, is applied after two-thirds of the total iterations, when most of the scene geometry has been reconstructed. More details can be found in the supplementary materials.
We train 10k iterations in 12-view settings and 30k iterations in 24-view settings. The depth correlation loss is applied with a patch size of $64 \times 64$ pixels for the MipNeRF360 dataset and $32 \times 32$ pixels for other datasets. The $\beta$ for the softmax depth is set to 5. Floater pruning is applied once after training completes. In general, the depth correlation loss converges after 3k iterations. Meanwhile, alpha-blending depth maps are rendered for image re-projection. For each training view, we generate four extra warping viewpoints by rotating the camera around the center of scene. 

We use Stable-Diffusion version 2.1 for our diffusion model in ~\cref{sec:diffusion_loss}. The SDS loss is enabled after two-thirds of the total iterations, when the majority of scene structure has been stable.

We set the depth warping loss weight to $0.05$, the SDS loss weight to $5 \times 10^{-4}$, the local and global Pearson loss weights both to $0.15$.

\subsection{Comparison}
\vspace{2pt}\noindent\textbf{Unbounded Dataset.}
\label{sec:mipnerf360}
We use the Mip-NeRF360 dataset to evaluate 3D reconstruction of unbounded 360\textdegree~scenes. The dataset presents significant challenges due to inherent complexity of scenes and strong occlusions under sparse-view settings. Moreover, the initial point cloud from COLMAP can be very limited even with the MVS step. In order to prove robustness of our method, we also evaluate performance with even sparser point clouds output by Structure From Motion (i.e., without the MVS step) as initialization. In quantitative evaluation from now on, we compute peak signal-to-noise ratio (PSNR), Learned Perceptual Image Patch Similarity (LPIPS) and structural similarity index measure (SSIM) metrics. As shown in~\cref{tab:result}, SparseGS significantly outperforms previous NeRF-based methods and concurrent works, FSGS and DNGaussian, in both 12-view and 24-view settings.

\begin{figure*}[tb]
  \centering
  \captionsetup[subfloat]{labelformat=empty, skip=5pt}
  {\includegraphics[width=0.95\widthqualFigure]{figures/qual_fig.jpg}}
  \caption{\textbf{Qualitative evaluation on the Mip-NeRF 360 dataset.} Our model effectively reconstructs high-frequency geometry, preserving sharp boundaries between the subject and the background. In contrast, FSGS excels in preserving fine details due to its densification technique but fails to reconstruct background geometry. On the other hand, DNGaussian can render a coherent background, but foreground objects are incorrectly pruned out.}
  \label{fig:qual_results}
  % \vspace{-0.1cm}
\end{figure*}

\begin{table}
\setlength{\abovecaptionskip}{4pt}
\resizebox{1\linewidth}{!}{
\setlength{\tabcolsep}{3.2 mm}
\begin{tabular}{l|ccc|ccc}
\toprule
& \multicolumn{3}{c|}{12-view} & \multicolumn{3}{c}{24-view} \\
 Models & PSNR $\uparrow$ & LPIPS $\downarrow$ & SSIM $\uparrow$ & PSNR $\uparrow$ & LPIPS $\downarrow$ & SSIM $\uparrow$ \\ \midrule

Mip-NeRF 360 & 17.73 & 0.520 & 0.432 & 19.78 & 0.431 & 0.530 \\
RegNeRF & \cellcolor[HTML]{FFFFD4}18.84 & 0.544 & 0.437 & 20.55 & 0.398 & 0.546 \\
SparseNeRF & 17.44 & 0.609 & 0.395 & 21.13 & 0.389 & 0.600 \\
\midrule 
3DGS & 17.49 & \cellcolor[HTML]{FFFFD4}0.431 & 0.499 & 19.93 & 0.401 & 0.588 \\
DNGaussian & 16.28 & 0.549 & 0.432 & 19.26 & 0.440 & 0.550 \\ 
\midrule 
FSGS (w/o MVS) & 18.15 & 0.485 & 0.504 & 21.76 & 0.396  & 0.626 \\

SparseGS (w/o MVS) & 18.46 & 0.476 & \cellcolor[HTML]{FFFFD4}0.513 & \cellcolor[HTML]{FFFFD4}22.03 & \cellcolor[HTML]{FFFFD4}0.381 & \cellcolor[HTML]{FFFFD4}0.631 \\

FSGS & \cellcolor[HTML]{FFE4CF}19.31 & \cellcolor[HTML]{FFE4CF}0.413 & \cellcolor[HTML]{FFE4CF}0.574 & \cellcolor[HTML]{FFE4CF}22.82 & \cellcolor[HTML]{FFE4CF}0.293 & \cellcolor[HTML]{FFE4CF}0.693 \\

\textbf{SparseGS (Ours)} & \cellcolor[HTML]{FFCCC9}19.37 & \cellcolor[HTML]{FFCCC9}0.398 & \cellcolor[HTML]{FFCCC9}0.577 & \cellcolor[HTML]{FFCCC9}23.02 &  \cellcolor[HTML]{FFCCC9}0.290 & \cellcolor[HTML]{FFCCC9}0.713 \\ \bottomrule
\end{tabular}
}
\caption{\textbf{Quantitative comparison on Mip-NeRF360 dataset for 12/24 input-view settings.}}
\vspace{-0.5cm}
\label{tab:result}
\end{table}

% \begin{figure}[tb]
%   \centering
%   {\includegraphics[width={0.47\textwidth}]{figures/pointcloud.png}}
  
%     \caption{\textbf{Both SFM and MVS methods struggle to generate high-quality point clouds when views are limited.} The point clouds shown here were produced using 12 input views on the Garden scene. The quality of these initial point clouds can significantly impact the reconstruction performance of 3DGS-based methods. See~\cref{tab:result} for comparisons.}
%     \label{fig:ptcomp}
%     \vspace{-0.5cm}
% \end{figure}

\vspace{2pt}\noindent\textbf{Forward-facing Datasets.}
\label{sec:forward_result}
 We also provide evaluations on the forward-facing datasets (LLFF and DTU) to demonstrate robustness of our pipeline. The LLFF dataset comprises eight complex forward-facing real scenes, while the DTU dataset includes object-centric scenes with foreground masks. It is important to note that in under-reconstructed regions, where input coverage is insufficient, NeRF-based methods often produce overly smooth appearances. This limitation actually prompted the introduction of positional encoding~\cite{mildenhall2020nerf, tancik2020fourier}. In contrast, 3DGS-based representations gravitate towards placing isolated Gaussians, appearing as high-frequency artifacts. PSNR tends to tolerate overly smooth images and heavily penalizes sharp artifacts, while perceptual metrics like LPIPS emphasize the opposite~\cite{1284395}. This can lead to minor discrepancies between the two metrics on certain test cases in~\cref{tab:forward_result}. In general, SparseGS is  competitive against previous and concurrent SOTA methods in both datasets.

\begin{table}[!t]
\setlength{\abovecaptionskip}{4pt}
\resizebox{1\linewidth}{!}{
\setlength{\tabcolsep}{3.2 mm}
\begin{tabular}{l|ccc|ccc}
\toprule
 & \multicolumn{3}{c|}{LLFF} & \multicolumn{3}{c}{DTU} \\
 Models & PSNR $\uparrow$ & LPIPS $\downarrow$ & SSIM $\uparrow$ & PSNR $\uparrow$ & LPIPS $\downarrow$ & SSIM $\uparrow$ \\ \midrule

Mip-NeRF & 15.22 & 0.540 & 0.351 & 16.71 & 0.239 & 0.757 \\
RegNeRF & 18.06 & 0.411 & 0.535 & \cellcolor[HTML]{FFFFD4}18.89 & \cellcolor[HTML]{FFFFD4}0.190 & 0.745 \\
SparseNeRF & 19.07 & 0.401 & 0.564 & \cellcolor[HTML]{FFCCC9}19.55 & 0.201 & \cellcolor[HTML]{FFFFD4}0.769 \\
\midrule 
3DGS & 16.94 & 0.402 & 0.488 & 14.18 & 0.301 & 0.628 \\
FSGS & \cellcolor[HTML]{FFCCC9}19.88 & \cellcolor[HTML]{FFFFD4}0.340 & \cellcolor[HTML]{FFE4CF}0.612 & 18.36 & 0.232 & 0.707 \\ 
DNGaussian & \cellcolor[HTML]{FFFFD4}19.12 & \cellcolor[HTML]{FFCCC9}0.294 & \cellcolor[HTML]{FFFFD4}0.591 & \cellcolor[HTML]{FFE4CF}18.91 & \cellcolor[HTML]{FFCCC9}0.176 & \cellcolor[HTML]{FFE4CF}0.790 \\ 
\textbf{SparseGS (Ours)} & \cellcolor[HTML]{FFE4CF}19.86 & \cellcolor[HTML]{FFE4CF}0.322 & \cellcolor[HTML]{FFCCC9}0.668 & \cellcolor[HTML]{FFFFD4}18.89 & \cellcolor[HTML]{FFE4CF}0.178 & \cellcolor[HTML]{FFCCC9}0.834 \\ \bottomrule
\end{tabular}
}
\caption{\textbf{Quantitative comparison on forward-facing datasets.}}
\vspace{-0.1cm}
\label{tab:forward_result}
\end{table}

\subsection{Ablation Studies}
\begin{figure*}[tb]
  \centering
  \captionsetup[subfloat]{labelformat=empty, skip=5pt}
  {\includegraphics[width=0.95\widthqualFigure]{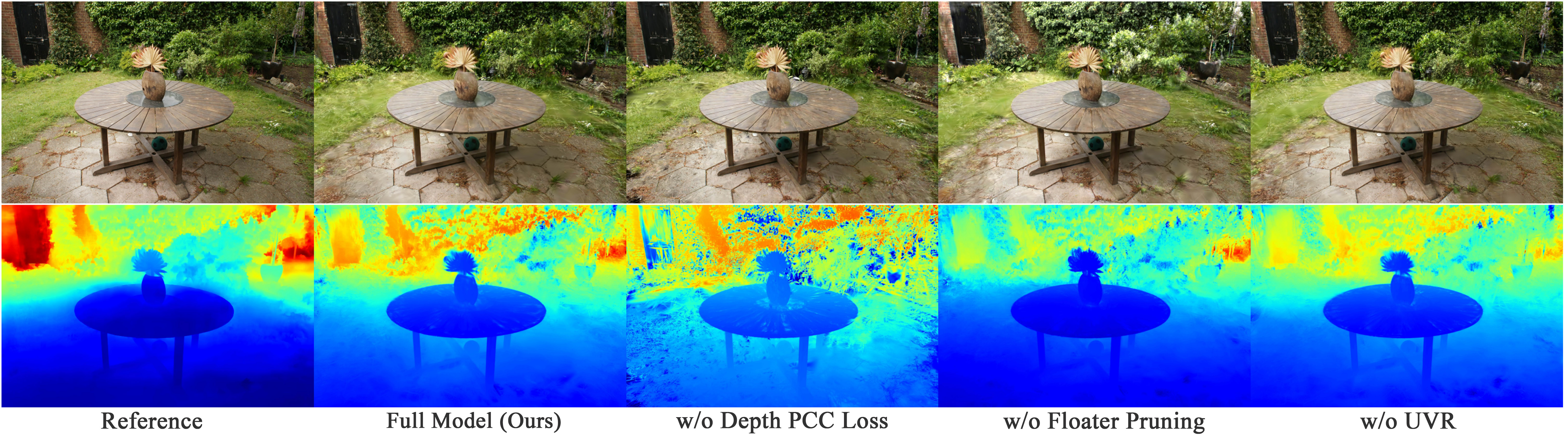}}
  \caption{\textbf{Ablation studies.} The reference depth map is produced by a monocular depth estimation model. Our complete model outputs a cleaner and more consistent scene structure with less noise.}
  \label{fig:ablation}
  \vspace{-0.5cm}
\end{figure*}
\begin{table}[tb]
\setlength{\tabcolsep}{3.5 mm}
\resizebox{1\linewidth}{!}{
% \begin{tabular}{cc|cc|ccc}
% \toprule
% \multicolumn{2}{c|}{Depth PCC}  & \multicolumn{2}{|c}{} & \multirow{2}{*}{PSNR $\uparrow$} & \multirow{2}{*}{LPIPS $\downarrow$} & \multirow{2}{*}{SSIM $\uparrow$} \\
% Alpha-blending & \multicolumn{1}{c|}{Softmax-scaling} & UVR  &  Pruning           &  & &                       
%   \\ \midrule
%     & \multicolumn{1}{c|}{}          &                 &                  
%   & 17.49                 & 0.431                  & 0.499                
%   \\ \midrule
%   \checkmark  & \multicolumn{1}{c|}{}      &       &                  
%   & 18.45                 & 0.410                  & 0.549                 
%   \\ 
%   & \multicolumn{1}{c|}{\checkmark} &              &     
%    & 18.86                & 0.401                  & 0.561                        
%   \\ \midrule
%   & \multicolumn{1}{c|}{\checkmark} & \checkmark   &                 
%   & 19.09                 & 0.400                  & 0.565                 
%   \\
%   & \multicolumn{1}{c|}{\checkmark} & \checkmark   & \checkmark                 
%   & \textbf{19.37}                 & \textbf{0.398}               & \textbf{0.577}     \\ \bottomrule
% \end{tabular}
\begin{tabular}{cc|cc|ccc}
\toprule
\multicolumn{2}{c|}{Depth PCC} & \multicolumn{2}{c|}{} & \multirow{2}{*}{PSNR $\uparrow$} & \multirow{2}{*}{LPIPS $\downarrow$} & \multirow{2}{*}{SSIM $\uparrow$} \\
Alpha-blending & Softmax-scaling & UVR & Pruning &  &  &  \\ 
\midrule
&  &  &  & 17.49 & 0.431 & 0.499 \\ 
\midrule
\checkmark &  &  &  & 18.45 & 0.410 & 0.549 \\ 
& \checkmark &  &  & 18.86 & 0.401 & 0.561 \\ 
\midrule
& \checkmark & \checkmark &  & 19.09 & 0.400 & 0.565 \\ 
& \checkmark & \checkmark & \checkmark & \textbf{19.37} & \textbf{0.398} & \textbf{0.577} \\ 
\bottomrule
\end{tabular}
}
\caption{\textbf{Ablation Studies.} We ablate our components on the Mip-NeRF360 dataset under 12-view setting.}
\vspace{-0.5cm}
\label{tab:ablation}
\end{table}

\label{sec:ablations}
\begin{figure}
  \centering
  \captionsetup[subfloat]{labelformat=empty, skip=5pt}
  {\includegraphics[width=0.46\widthqualFigure]{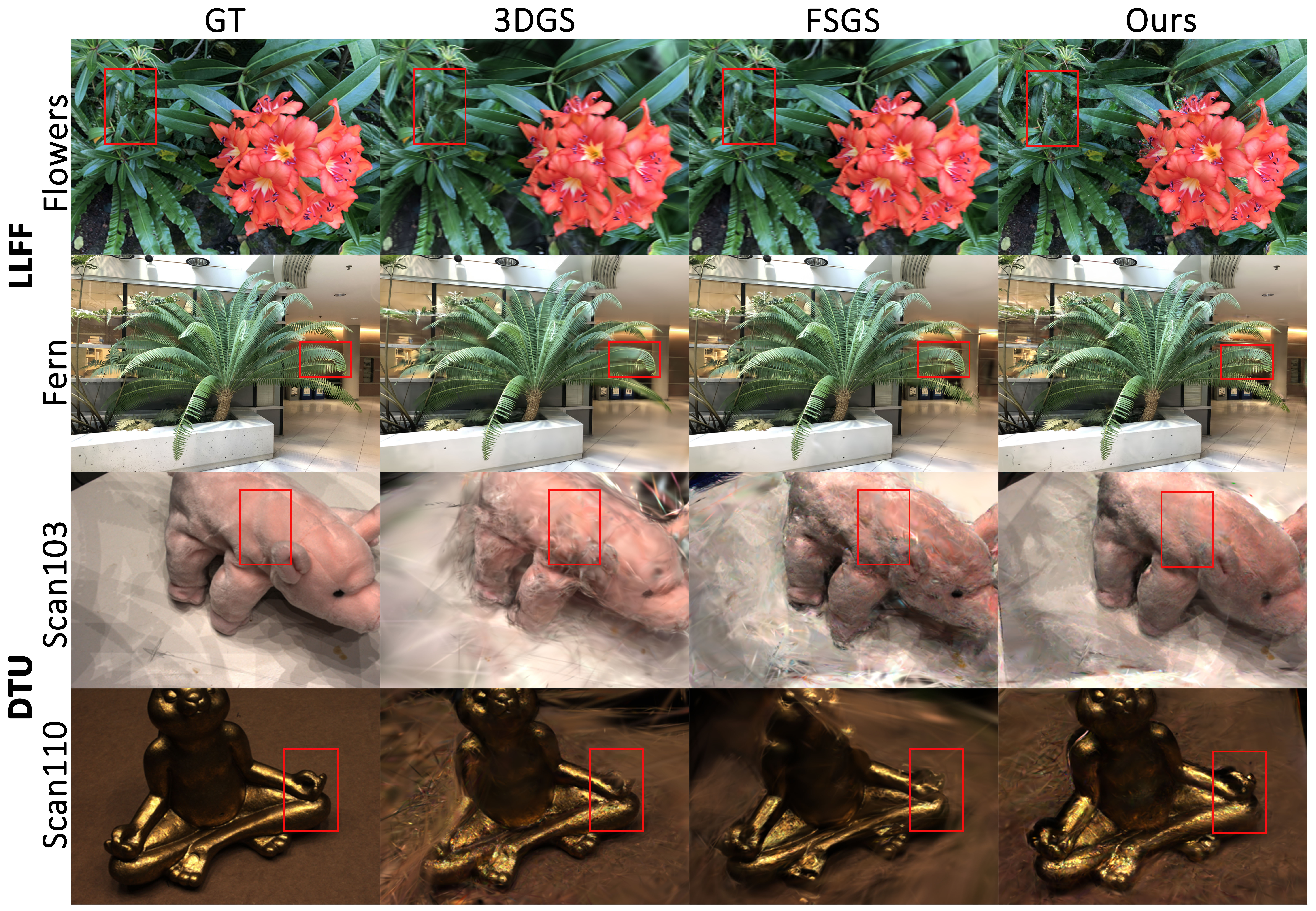}}
  \caption{\textbf{Qualitative evaluation on the forward-facing datasets.} Our method is able to reconstruct high-frequency geometry more accurately than state-of-the-art~\cite{zhu2023FSGS}.}
  \label{fig:llff_qual_results}
  \vspace{-0.5cm}
\end{figure}

We ablate our method on the Mip-NeRF360 dataset 12-view setting. Quantitative results are reported in~\cref{tab:ablation}. Qualitative figures are shown in~\cref{fig:ablation}.

\vspace{2pt}\noindent\textbf{Depth Correlation Loss.} The depth Pearson Correlation loss introduces depth knowledge into the 3DGS representation. We show the improvements by applying this loss to both alpha-blending and softmax-scaling depth. As indicated in~\cref{tab:ablation}, our softmax-scaling depth rendering method performs better, improving PSNR by 1.37dB compared to 3DGS and significantly enhance the quality of the rendered depth map~\cref{fig:ablation}.

\vspace{2pt}\noindent\textbf{Unseen Viewpoints Regularization.} The UVR module adds regularization from viewpoints away from the training viewpoints, reducing the problem of excessive overfitting in sparse-input settings. It effectively reduces high-frequency artifacts, improving PSNR by an average 0.33dB.

\vspace{2pt}\noindent\textbf{Floater Pruning.} The floater pruning heuristic further cleans up the scene by deleting misplaced low-opacity Gaussians, making both the image and the depth map sharper. This module further boosts PSNR by 0.28dB. 

%% file: sec/5_conclusion.tex
\section{Conclusion}
\label{sec:conclusion}
% In this paper, we introduce a novel 3DGS-based technique for few-shot novel view synthesis. We leverage the explicit nature of the 3D gaussian representation to introduce a novel pruning operator designed to reduce and remove "floater" artifacts. Applied to the Mip-NeRF 360 dataset, we show our technique can achieve state-of-the-art performance on few shot novel view synthesis for 360\textdegree~unbounded scenes.

In this paper, we propose a method using 3D Gaussian Splatting (3DGS) representation to tackle sparse-view 3D reconstruction task. We observe that the alpha blending depth rendering in original 3DGS often results in misplaced Gaussians (known as ``floaters''). Therefore, we propose to constrain depth convergence with softmax-scaling and mode-selection bias that significantly reduce such floaters. In regions with little coverage by input views, we leverage Score Distillation Sampling (SDS) and Depth Warping to reduce collapse in geometry and noise in texture while preserving fine details. Lastly, we propose a novel floater pruning process to identify and remove low-opacity floaters. In evaluation, we show that our method outperforms the state-of-the-art methods by outputting a much cleaner and more coherent scene under even more challenging 12-view setting on Mip-NeRF360 dataset.

%% file: sec/X_supp_new.tex
\definecolor{bg}{gray}{0.95}

\crefname{section}{Sec.}{Secs.}
\Crefname{section}{Section}{Sections}
\Crefname{table}{Table}{Tables}
\crefname{table}{Tab.}{Tabs.}

\renewcommand\thesection{\Alph{section}} % section with alphabet
\renewcommand\thesubsection{\thesection.\alph{subsection}} % subsection with alphabet
\renewcommand\thefigure{\Alph{figure}} % figure with alphabet
\renewcommand\thetable{\Alph{table}} % table with alphabet

\newcommand{\dmode}[1]{$d^{\text{mode}}$}

\clearpage
\onecolumn
{
    \centering
    \Large
    \textbf{\thetitle}\\
    \vspace{0.5em}Supplementary Material \\
    \vspace{1.0em}
}

\section*{Supplementary Contents}
This supplementary is organized as follows:
\begin{itemize}
    \item \cref{sec:soft_derivation} presents the gradient derivation for our softmax-scaling depth formulation.
    \item \cref{sec:view_sampling} details the view sampling methods used in our UVR module.
    \item \cref{sec:floater_prune} provides algorithm details for our Advance Floater Pruning method.
\end{itemize}

\section{Derivation of the Softmax Depth Gradient}
\label{sec:soft_derivation}
In this section, we derive the gradients used for the softmax depth implemented in the rasterizer. We assume we are rendering a single pixel and that the Gaussians $G_i$ along the camera ray are ordered from closest to the camera plane to the farthest. (i.e. $\alpha_1$ corresponds to the Gaussian closest to the camera, while $\alpha_N$ corresponds to the farthest)\\
\noindent
If we let $w_k = T_k \alpha_k$ and $T_k = \prod_{i=1}^{k-1} (1- \alpha_i)$\cite{kerbl3Dgaussians}, then

\begin{equation}
    d^{\text{softmax}}_{x,y} = \log \left( \frac{ \sum_{i=1}^N w_i e^{ \beta w_i}  d_i}{\sum_{i=1}^N w_i e^{\beta w_i} } \right) 
\end{equation}

The derivative w.r.t the Gaussian's camera depth value is:

\begin{equation}
     \frac{\partial d^{\text{softmax}}_{x,y}}{\partial d_k} = \frac{w_k e^{ \beta w_k}}{ \sum_{i=1}^N w_i e^{ \beta w_i}  d_i} 
\end{equation}

The derivative w.r.t the Gaussian's alpha value is:

    \begin{align*}
         &\frac{\partial d^{\text{softmax}}_{x,y}}{\partial \alpha_k} = \\&\qquad\qquad \frac{ (1 + \beta w_k) T_k e ^ {\beta w_k} d_k  - \frac{1}{1 - \alpha_k} \sum_{j = k + 1}^ N 
 (1 + \beta w_j) w_j e^ {\beta w_j} d_j  }{  \sum_{i=1}^N w_i e^{ \beta w_i}  d_i } \tag{3} \\  &\qquad\qquad- \frac{ (1 + \beta w_k) T_k e ^ {\beta w_k}   - \frac{1}{1 - \alpha_k} \sum_{j = k + 1}^ N 
 (1 + \beta w_j) w_j e^ {\beta w_j}  }{  \sum_{i=1}^N w_i e^{ \beta w_i}  } \tag{4}
    \end{align*}

The sums in the denominators are retained during the forward pass and subsequently transferred to the backward pass. To avoid an $O(n^2)$ blowup, an accumulator strategy is used as in \cite{kerbl3Dgaussians}. 

Let $\mathcal{A}^l_k,\mathcal{A}^r_k$  be the accumulators pertaining to $\frac{\partial d^{\text{softmax}}_{x,y}}{\partial \alpha_k}$ for expressions (3) and (4) respectively. 
\begin{align*}
        \mathcal{A}^l_0 &= \mathcal{A}^r_0 = 0 \\
        \mathcal{A}^l_k &= \alpha_{k-1} (1 + \beta w_{k-1}) e^{\beta w_{k-1}}  d_{k-1} + (1 - \alpha_{k-1}) \mathcal{A}^l_{k-1} \\
        \mathcal{A}^r_k &= \alpha_{k-1} (1 + \beta w_{k-1}) e^{\beta w_{k-1}} + (1 - \alpha_{k-1}) \mathcal{A}^r_{k-1}  
\end{align*}

It can be shown that

\begin{equation*}
\frac{\partial d^{\text{softmax}}_{x,y}}{\partial \alpha_k} = T_k \left  ( \frac{(1 + \beta w_{k}) e^{\beta w_{k}} d_k  - \mathcal{A}^l_k}{ \sum_{i=1}^N w_i e^{ \beta w_i}  d_i} - \frac{(1 + \beta w_{k}) e^{\beta w_{k}} - \mathcal{A}^r_k}{ \sum_{i=1}^N w_i e^{ \beta w_i}} \right )
\end{equation*}

\newpage
\section{View Sampling for UVR}
\label{sec:view_sampling}
As described in the main paper, although the UVR as a whole is designed to regularize 3DGS training from sampled views, each component has a different strategy of camera sampling. 

\vspace{2pt}\noindent\textbf{Elliptic Cylinder Sampled Views.} To compute the SDS loss, we generate random camera views directed at the scene center using an elliptical cylinder that best fits the positions of the training views. The process begins by transforming the camera positions to align with the world coordinate system (where the positive Z-axis is perpendicular to the ground) using PCA. We then determine the scaling factors for the elliptical axes by taking the 90th percentile of the absolute displacements from the scene center, ensuring the ellipse encompasses most of the input views.

The vertical bounds for the z-coordinate are set at the 10th and 90th percentiles of the input views' z-coordinates. We uniformly randomly select a camera translation vector with x-y coordinates on the ellipse and a z-coordinate within the aforementioned z-bounds. Finally, the camera's rotation matrix is computed by directing its look-at vector towards the scene center and aligning the up vector with the average up vector of the input views.

\begin{figure}[h]
    \centering
    \includegraphics[width=0.4\textwidth]{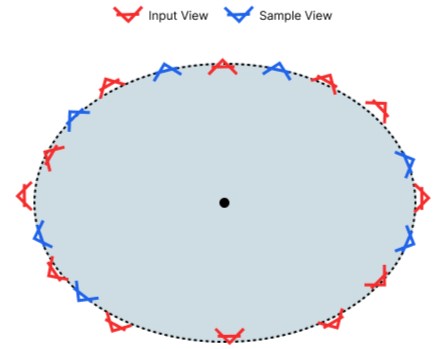}
    \caption{\textbf{Bird's Eye View of Elliptic Cylinder sampling.} The dashed line is the ellipse best fit to the input views (in red). The blue views are the sampled views on the ellipse which also have a random variation in the z-coordinate.}
    \label{fig:ellipsesampling}
\end{figure}

\vspace{2pt}\noindent\textbf{Warped Views.} We cannot use the same sampled viewpoints as in SDS because they are too far to provide large view overlapping. Instead, for each training camera, we apply a warp by a predefined angle, $\theta$, around the average up-vector of all the training cameras. Specifically, the up-vector of each camera is extracted as:
\begin{align}
   Y = \begin{bmatrix} R_{11} \\ R_{21} \\ R_{31} \end{bmatrix}
\end{align} with $R$ denoting the rotation matrix in the camera pose. Next, we compute the normalized average up-vector across all cameras to serve as an estimate of the scene center and orientation. Novel viewpoints are then created by rotating the training cameras around this estimated center axis by various angles using the Rodrigues' rotation formula~\cite{rodrigues1840lois}. 
\begin{gather}
\Bar{Y} = \frac{1}{N} \sum_i Y_i\\
\Bar{Y} = \frac{\Bar{Y}}{||\Bar{Y}||}\\
P' = P \cos{\theta} + (P \times \Bar{Y}) \sin{\theta} + \Bar{Y} (\Bar{Y} \cdot P)(1 - \cos{\theta})
\end{gather}
where $P$ represents the original camera pose, and $\Bar{Y}$ is the estimated up-axis at the scene center. We select the angles $\theta$ to be -3\textdegree,-1.5\textdegree, 1.5\textdegree, and 3\textdegree, resulting in four warped viewpoints for each training camera.

\begin{table*}
\centering
\begin{tabular}{c|c|c|c|c}
\toprule
  Apply at & Pearson Depth Loss & SDS Loss & Depth Warping Loss  & Floater Pruning
  \\ \midrule
  \multicolumn{1}{c|}{Input Views (12/24)}  & \multicolumn{1}{c|}{\checkmark}  & \multicolumn{1}{c|}{}    &   \multicolumn{1}{c|}{}
  &     \checkmark              
  \\ \midrule
   \multicolumn{1}{c|}{Elliptic Cylinder Sampled Views} &   \multicolumn{1}{c|}{}    &   \multicolumn{1}{c|}{\checkmark}    &        \multicolumn{1}{c|}{}          
  &                     
  \\ \midrule
 \multicolumn{1}{c|}{Warped Views} & \multicolumn{1}{c|}{} &   \multicolumn{1}{c|}{}           &   \multicolumn{1}{c|}{\checkmark}  
   &               
 \\ \bottomrule
\end{tabular}

\caption{The Pearson Depth loss and floater pruning are applied to sparse input views, whereas the SDS and Depth Warping losses are applied to differently sampled views.}
\label{tab:sampling}
\end{table*}

\newpage
\section{Advance Floater Pruning Algorithm}
\label{sec:floater_prune}
In 3DGS, ``floaters'' refer to relatively low-opacity Gaussians positioned close to the camera center. These floaters do not appear prominently when rendering the alpha-blending depth of a scene because they are ``averaged out'', but they stand out in the mode-selection depth. Since the Pearson depth loss is a relatively a soft constraint, misalignments between the mode and alpha depth can still occur. This misalignment is largely due to 3DGS halting rasterization once a transmittance threshold is reached. The floater Gaussians near the camera deplete this transmittance budget early, preventing other Gaussians from converging at the correct depth.

To address this, we threshold the areas of misalignment to generate a floater mask $F$ for each input sparse view to identify and delete Gaussians that are incorrectly placed too close to the camera. Specifically, we first compute the relative difference between the mode-selection depth and alpha-blending depth for each training view $i$ as $\Delta_i = \frac{(d^{mode}-d^{alpha})}{d^{alpha}}$. As discussed in the main paper, when visualizing the distributions of $\Delta_i$, images with many floaters have bimodal histograms, as floaters tend to deviate significantly from the true depth. Based on this observation, we apply dip test~\cite{10.1214/aos/1176346577}, a measure of uni-modality, on the distribution. We then average the uni-modality score across all training views for a scene, since the presence of floaters is a scene-wide metric, and use the mean score to select a percentile threshold for the relative differences.

The conversion from the dip statistic to a threshold is done using an exponential curve with parameters $a$ and $b$. We estimate these parameters by manually examining $\Delta_i$ and $F_i$ for various scenes from different datasets and real world capture, and our values are $a = 0.97, b = -7.5$. This process was carefully designed to allow our floater pruning method to be adaptive. The full pruning algorithm is as follow: 

% \SetKwProg{Fn}{Function}{}{end}
% \renewcommand{\algorithmcfname}{ALGORITHM}
% \SetAlFnt{\small}
% \SetAlCapFnt{\small}
% \SetAlCapNameFnt{\small}
% \SetAlCapHSkip{0pt}

\begin{algorithm}
\caption{Algorithm to Prune Floaters}
\label{algo:floaterprune}
\begin{algorithmic}[1]
\State \textbf{Input:} 3DGS Representation $G$, Training Cameras $I$, Curve Parameters $a$, $b$
\State $\Bar{D} \gets 0.0$
\ForAll{$i \in I$}
    \State $d^{\text{alpha}} \gets \text{alpha\_blending\_depth}(G, i)$
    \State $d^{\text{mode}} \gets \text{mode\_selection}(G, i)$
    \State $\Delta_i \gets \frac{d^{\text{mode}} - d^{\text{alpha}}}{d^{\text{alpha}}}$
    \State $\Bar{D} \gets \Bar{D} + \text{dip\_test}(\Delta_i)$
\EndFor
\State $\Bar{D} \gets \Bar{D} / |I|$
\ForAll{$i \in I$}
    \State $\tau_i \gets \text{percentile}(\Delta_i, a e^{b \Bar{D}})$
    \State $F_i \gets \mathbb{1}[\Delta_i > \tau_i]$
    \State $\{g_0, g_1, ..., g_n\} \gets \text{mask\_to\_Gaussian}(F_i)$
    \State \text{remove\_Gaussians}($\{g_0, g_1, ..., g_n\}$)
\EndFor
\end{algorithmic}
\end{algorithm}

\newpage
\section{More Qualitative Comparisons with Few-shot Methods~\cite{zhu2023FSGS,wang2023sparsenerf}}
\vspace{-0.3cm}
\begin{figure}[h]
    \centering
    \includegraphics[width=0.94\textwidth]{figures/suppl.jpg}
    \caption{\textbf{Qualitative comparisons with FSGS and SparseNeRF on the MipNeRF360 dataset.} All experiments above are trained with 12 input views. SparseNeRF often over-smooths regions that lack coverage from the training views. In contrast, our method preserves sharp details and more accurately fills in these missing regions. FSGS excels in maintaining foreground structure and details but can lack background coherence and produce floaters.}
    \label{fig:suppl}
    \vspace{-0.5cm}
\end{figure}

%% file: main.bib
@String(CVPR= {IEEE Conf. Comput. Vis. Pattern Recog.})

@String(ICCV= {Int. Conf. Comput. Vis.})

@String(ECCV= {Eur. Conf. Comput. Vis.})

@String(TOG= {ACM Trans. Graph.})

@String(CVPR  = {CVPR})

@String(ICCV  = {ICCV})

@String(ECCV  = {ECCV})

@String(TOG   = {ACM TOG})

@Article{kerbl3Dgaussians,
      author       = {Kerbl, Bernhard and Kopanas, Georgios and Leimk{\"u}hler, Thomas and Drettakis, George},
      title        = {3D Gaussian Splatting for Real-Time Radiance Field Rendering},
      journal      = {ACM Transactions on Graphics},
      number       = {4},
      volume       = {42},
      month        = {July},
      year         = {2023},
      url          = {https://repo-sam.inria.fr/fungraph/3d-gaussian-splatting/}
}

@inproceedings{mildenhall2020nerf,
  title={NeRF: Representing Scenes as Neural Radiance Fields for View Synthesis},
  author={Ben Mildenhall and Pratul P. Srinivasan and Matthew Tancik and Jonathan T. Barron and Ravi Ramamoorthi and Ren Ng},
  year={2020},
  booktitle={ECCV},
}

@InProceedings{kangle2021dsnerf,
    author    = {Deng, Kangle and Liu, Andrew and Zhu, Jun-Yan and Ramanan, Deva},
    title     = {Depth-supervised {NeRF}: Fewer Views and Faster Training for Free},
    booktitle = {Proceedings of the IEEE/CVF Conference on Computer Vision and Pattern Recognition (CVPR)},
    month     = {June},
    year      = {2022}
}

@article{Xu_2022_neuralLift,
title = {NeuralLift-360: Lifting An In-the-wild 2D Photo to A 3D Object with 360° Views},
author = {Xu, Dejia and Jiang, Yifan and Wang, Peihao and Fan, Zhiwen and Wang, Yi and Wang, Zhangyang},
journal={arXiv preprint arXiv:2211.16431},
year={2022}
}

@article{tang2023dreamgaussian,
  title={Dreamgaussian: Generative gaussian splatting for efficient 3d content creation},
  author={Tang, Jiaxiang and Ren, Jiawei and Zhou, Hang and Liu, Ziwei and Zeng, Gang},
  journal={arXiv preprint arXiv:2309.16653},
  year={2023}
}

@inproceedings{jensen2014large,
  title={Large scale multi-view stereopsis evaluation},
  author={Jensen, Rasmus and Dahl, Anders and Vogiatzis, George and Tola, Engil and Aan{\ae}s, Henrik},
  booktitle={2014 IEEE Conference on Computer Vision and Pattern Recognition},
  pages={406--413},
  year={2014},
  organization={IEEE}
}

@article{liu2023one,
  title={One-2-3-45: Any single image to 3d mesh in 45 seconds without per-shape optimization},
  author={Liu, Minghua and Xu, Chao and Jin, Haian and Chen, Linghao and Xu, Zexiang and Su, Hao and others},
  journal={arXiv preprint arXiv:2306.16928},
  year={2023}
}

@article{tancik2020fourier,
  title={Fourier features let networks learn high frequency functions in low dimensional domains},
  author={Tancik, Matthew and Srinivasan, Pratul and Mildenhall, Ben and Fridovich-Keil, Sara and Raghavan, Nithin and Singhal, Utkarsh and Ramamoorthi, Ravi and Barron, Jonathan and Ng, Ren},
  journal={Advances in Neural Information Processing Systems},
  volume={33},
  pages={7537--7547},
  year={2020}
}

@article{Ranftl2020,
	author    = {Ren\'{e} Ranftl and Katrin Lasinger and David Hafner and Konrad Schindler and Vladlen Koltun},
	title     = {Towards Robust Monocular Depth Estimation: Mixing Datasets for Zero-shot Cross-dataset Transfer},
	journal   = {IEEE Transactions on Pattern Analysis and Machine Intelligence (TPAMI)},
	year      = {2020},
}

@article{Ranftl2021,
	author    = {Ren\'{e} Ranftl and Alexey Bochkovskiy and Vladlen Koltun},
	title     = {Vision Transformers for Dense Prediction},
	journal   = {ArXiv preprint},
	year      = {2021},
}

@inproceedings{pix3d,
  title={Pix3D: Dataset and Methods for Single-Image 3D Shape Modeling},
  author={Sun, Xingyuan and Wu, Jiajun and Zhang, Xiuming and Zhang, Zhoutong and Zhang, Chengkai and Xue, Tianfan and Tenenbaum, Joshua B and Freeman, William T},
  booktitle={IEEE Conference on Computer Vision and Pattern Recognition (CVPR)},
  year={2018}
}

@article{hong2023lrm,
  title={LRM: Large Reconstruction Model for Single Image to 3D},
  author={Hong, Yicong and Zhang, Kai and Gu, Jiuxiang and Bi, Sai and Zhou, Yang and Liu, Difan and Liu, Feng and Sunkavalli, Kalyan and Bui, Trung and Tan, Hao},
  journal={arXiv preprint arXiv:2311.04400},
  year={2023}
}

@INPROCEEDINGS{4547850,
  author={Mori, Yuji and Fukushima, Norishige and Fujii, Toshiaki and Tanimoto, Masayuki},
  booktitle={2008 3DTV Conference: The True Vision - Capture, Transmission and Display of 3D Video}, 
  title={View Generation with 3D Warping Using Depth Information for FTV}, 
  year={2008},
  volume={},
  number={},
  pages={229-232},
  doi={10.1109/3DTV.2008.4547850}}

@InProceedings{ke2023repurposing,
      title={Repurposing Diffusion-Based Image Generators for Monocular Depth Estimation},
      author={Bingxin Ke and Anton Obukhov and Shengyu Huang and Nando Metzger and Rodrigo Caye Daudt and Konrad Schindler},
      booktitle = {Proceedings of the IEEE/CVF Conference on Computer Vision and Pattern Recognition (CVPR)},
      year={2024}
}

@article{mildenhall2019llff,
  title={Local Light Field Fusion: Practical View Synthesis with Prescriptive Sampling Guidelines},
  author={Ben Mildenhall and Pratul P. Srinivasan and Rodrigo Ortiz-Cayon and Nima Khademi Kalantari and Ravi Ramamoorthi and Ren Ng and Abhishek Kar},
  journal={ACM Transactions on Graphics (TOG)},
  year={2019},
}

@misc{rombach2021highresolution,
      title={High-Resolution Image Synthesis with Latent Diffusion Models}, 
      author={Robin Rombach and Andreas Blattmann and Dominik Lorenz and Patrick Esser and Björn Ommer},
      year={2021},
      eprint={2112.10752},
      archivePrefix={arXiv},
      primaryClass={cs.CV}
}

@article{yi2023gaussiandreamer,
  title={GaussianDreamer: Fast Generation from Text to 3D Gaussian Splatting with Point Cloud Priors},
  author={Yi, Taoran and Fang, Jiemin and Wu, Guanjun and Xie, Lingxi and Zhang, Xiaopeng and Liu, Wenyu and Tian, Qi and Wang, Xinggang},
  journal={arXiv preprint arXiv:2310.08529},
  year={2023}
}

@article{li2024dngaussian,
   title={DNGaussian: Optimizing Sparse-View 3D Gaussian Radiance Fields with Global-Local Depth Normalization},
   author={Li, Jiahe and Zhang, Jiawei and Bai, Xiao and Zheng, Jin and Ning, Xin and Zhou, Jun and Gu, Lin},
   journal={arXiv preprint arXiv:2403.06912},
   year={2024}
}

@misc{liu2023zero1to3,
      title={Zero-1-to-3: Zero-shot One Image to 3D Object}, 
      author={Ruoshi Liu and Rundi Wu and Basile Van Hoorick and Pavel Tokmakov and Sergey Zakharov and Carl Vondrick},
      year={2023},
      eprint={2303.11328},
      archivePrefix={arXiv},
      primaryClass={cs.CV}
}

@article{stan2023ldm3d,
  title={LDM3D: Latent Diffusion Model for 3D},
  author={Stan, Gabriela Ben Melech and Wofk, Diana and Fox, Scottie and Redden, Alex and Saxton, Will and Yu, Jean and Aflalo, Estelle and Tseng, Shao-Yen and Nonato, Fabio and Muller, Matthias and others},
  journal={arXiv preprint arXiv:2305.10853},
  year={2023}
}

@misc{zhu2023FSGS, 
title={FSGS: Real-Time Few-Shot View Synthesis using Gaussian Splatting}, 
author={Zehao Zhu and Zhiwen Fan and Yifan Jiang and Zhangyang Wang}, 
year={2023},
eprint={2312.00451},
archivePrefix={arXiv},
primaryClass={cs.CV} 
}

@article{jain2021dreamfields,
  title = {Zero-Shot Text-Guided Object Generation with Dream Fields},
  author = {Jain, Ajay and Mildenhall, Ben and Barron, Jonathan T. and Abbeel, Pieter and Poole, Ben},
  journal = {CVPR},
  year = {2022},
}

@article{wu20234d,
  title={4d gaussian splatting for real-time dynamic scene rendering},
  author={Wu, Guanjun and Yi, Taoran and Fang, Jiemin and Xie, Lingxi and Zhang, Xiaopeng and Wei, Wei and Liu, Wenyu and Tian, Qi and Wang, Xinggang},
  journal={arXiv preprint arXiv:2310.08528},
  year={2023}
}

@article{zielonka2023drivable,
  title={Drivable 3D Gaussian Avatars},
  author={Zielonka, Wojciech and Bagautdinov, Timur and Saito, Shunsuke and Zollh{\"o}fer, Michael and Thies, Justus and Romero, Javier},
  journal={arXiv preprint arXiv:2311.08581},
  year={2023}
}

@article{luiten2023dynamic,
  title={Dynamic 3d gaussians: Tracking by persistent dynamic view synthesis},
  author={Luiten, Jonathon and Kopanas, Georgios and Leibe, Bastian and Ramanan, Deva},
  journal={arXiv preprint arXiv:2308.09713},
  year={2023}
}

@article{1284395,
  author={Zhou Wang and Bovik, A.C. and Sheikh, H.R. and Simoncelli, E.P.},
  journal={IEEE Transactions on Image Processing}, 
  title={Image quality assessment: from error visibility to structural similarity}, 
  year={2004},
  volume={13},
  number={4},
  pages={600-612},
  doi={10.1109/TIP.2003.819861}}

@inproceedings{irshad2023neo,
  title={NeO 360: Neural Fields for Sparse View Synthesis of Outdoor Scenes},
  author={Irshad, Muhammad Zubair and Zakharov, Sergey and Liu, Katherine and Guizilini, Vitor and Kollar, Thomas and Gaidon, Adrien and Kira, Zsolt and Ambrus, Rares},
  booktitle={Proceedings of the IEEE/CVF International Conference on Computer Vision},
  pages={9187--9198},
  year={2023}
}

@inproceedings{rombach2022high,
  title={High-resolution image synthesis with latent diffusion models},
  author={Rombach, Robin and Blattmann, Andreas and Lorenz, Dominik and Esser, Patrick and Ommer, Bj{\"o}rn},
  booktitle={Proceedings of the IEEE/CVF conference on computer vision and pattern recognition},
  pages={10684--10695},
  year={2022}
}

@article{barron2021mipnerf,
    title={Mip-NeRF: A Multiscale Representation 
           for Anti-Aliasing Neural Radiance Fields},
    author={Jonathan T. Barron and Ben Mildenhall and 
            Matthew Tancik and Peter Hedman and 
            Ricardo Martin-Brualla and Pratul P. Srinivasan},
    journal={ICCV},
    year={2021}
}

@inproceedings{Miangoleh2021Boosting,
  title     = {Boosting Monocular Depth Estimation Models to High-Resolution via Content-Adaptive Multi-Resolution Merging},
  author    = {S. Mahdi H. Miangoleh and Sebastian Dille and Long Mai and Sylvain Paris and Ya\u{g}{\i}z Aksoy},
  booktitle = {Proc. CVPR},
  year = {2021}
}

@article{rodrigues1840lois,
  title={Lois du mouvement des systèmes de points matériels},
  author={Rodrigues, O.},
  journal={Journal de l'Ecole Royale Polytechnique},
  volume={19},
  pages={319--429},
  year={1840}
}

@inproceedings{levis2022gravitationally,
 title={Gravitationally Lensed Black Hole Emission Tomography},
author={Levis, Aviad and Srinivasan, Pratul P and Chael, Andrew A and Ng, Ren and Bouman, Katherine L},
booktitle={Proceedings of the IEEE/CVF Conference on Computer Vision and Pattern Recognition},
pages={19841--19850},
year={2022}
}

@InProceedings{weng_humannerf_2022_cvpr,
    title     = {Human{N}e{RF}: Free-Viewpoint Rendering of Moving People From Monocular Video},
    author    = {Weng, Chung-Yi and 
                 Curless, Brian and 
                 Srinivasan, Pratul P. and 
                 Barron, Jonathan T. and 
                 Kemelmacher-Shlizerman, Ira},
    booktitle = {Proceedings of the IEEE/CVF Conference on Computer Vision and Pattern Recognition (CVPR)},
    month     = {June},
    year      = {2022},
    pages     = {16210-16220}
}

@article{zhang2021nerfactor,
  title={Nerfactor: Neural factorization of shape and reflectance under an unknown illumination},
  author={Zhang, Xiuming and Srinivasan, Pratul P and Deng, Boyang and Debevec, Paul and Freeman, William T and Barron, Jonathan T},
  journal={ACM Transactions on Graphics (ToG)},
  volume={40},
  number={6},
  pages={1--18},
  year={2021},
  publisher={ACM New York, NY, USA}
}

@article{srinivasan19,
  author    = {Pratul P. Srinivasan and Richard Tucker and Jonathan T. Barron and Ravi Ramamoorthi and Ren Ng and Noah Snavely},
  title     = {Pushing the Boundaries of View Extrapolation with Multiplane Images},
  journal   = {CVPR},
  year      = {2019}}

@article{mildenhall2022rawnerf,
    title={{NeRF} in the Dark: High Dynamic Range View Synthesis from Noisy Raw Images},
    author={Ben Mildenhall and Peter Hedman and Ricardo Martin-Brualla and Pratul P. Srinivasan and Jonathan T. Barron},
    journal={CVPR},
    year={2022}
}

@inproceedings{tucker2020single,
  title={Single-view view synthesis with multiplane images},
  author={Tucker, Richard and Snavely, Noah},
  booktitle={Proceedings of the IEEE/CVF Conference on Computer Vision and Pattern Recognition},
  pages={551--560},
  year={2020}
}

@article{zhou2018stereo,
  title={Stereo magnification: Learning view synthesis using multiplane images},
  author={Zhou, Tinghui and Tucker, Richard and Flynn, John and Fyffe, Graham and Snavely, Noah},
  journal={arXiv preprint arXiv:1805.09817},
  year={2018}
}

@InProceedings{Jain_2021_ICCV,
  author = {Jain, Ajay and Tancik, Matthew and Abbeel, Pieter},
  title = {Putting NeRF on a Diet: Semantically Consistent Few-Shot View Synthesis},
  booktitle = {Proceedings of the IEEE/CVF International Conference on Computer Vision (ICCV)},
  month = {October},
  year = {2021},
  pages = {5885-5894}
}

@article{Xu_2022_SinNeRF,
    title = {SinNeRF: Training Neural Radiance Fields on Complex Scenes from a Single Image},
    author = {Xu, Dejia and Jiang, Yifan and Wang, Peihao and Fan, Zhiwen and Shi, Humphrey and Wang, Zhangyang},
    journal={arXiv preprint arXiv:2204.00928},
    year={2022}
}

@inproceedings{schoenberger2016sfm,
    author={Sch\"{o}nberger, Johannes Lutz and Frahm, Jan-Michael},
    title={Structure-from-Motion Revisited},
    booktitle={Conference on Computer Vision and Pattern Recognition (CVPR)},
    year={2016},
}

@inproceedings{schoenberger2016mvs,
    author={Sch\"{o}nberger, Johannes Lutz and Zheng, Enliang and Pollefeys, Marc and Frahm, Jan-Michael},
    title={Pixelwise View Selection for Unstructured Multi-View Stereo},
    booktitle={European Conference on Computer Vision (ECCV)},
    year={2016},
}

@article{wang2023sparsenerf,
    title={SparseNeRF: Distilling Depth Ranking for Few-shot Novel View Synthesis},
    author={Guangcong Wang and Zhaoxi Chen and Chen Change Loy and Ziwei Liu},
    journal={IEEE/CVF International Conference on Computer Vision (ICCV)},
    year={2023}}

@InProceedings{Niemeyer2021Regnerf,
    author    = {Michael Niemeyer and Jonathan T. Barron and Ben Mildenhall and Mehdi S. M. Sajjadi and Andreas Geiger and Noha Radwan},  
    title     = {RegNeRF: Regularizing Neural Radiance Fields for View Synthesis from Sparse Inputs},
    booktitle = {Proc. IEEE Conf. on Computer Vision and Pattern Recognition (CVPR)},
    year      = {2022},
}

@article{poole2022dreamfusion,
  title={Dreamfusion: Text-to-3d using 2d diffusion},
  author={Poole, Ben and Jain, Ajay and Barron, Jonathan T and Mildenhall, Ben},
  journal={arXiv preprint arXiv:2209.14988},
  year={2022}
}

@article{ho2020denoising,
  title={Denoising diffusion probabilistic models},
  author={Ho, Jonathan and Jain, Ajay and Abbeel, Pieter},
  journal={Advances in neural information processing systems},
  volume={33},
  pages={6840--6851},
  year={2020}
}

@article{po2023state,
  title={State of the Art on Diffusion Models for Visual Computing},
  author={Po, Ryan and Yifan, Wang and Golyanik, Vladislav and Aberman, Kfir and Barron, Jonathan T and Bermano, Amit H and Chan, Eric Ryan and Dekel, Tali and Holynski, Aleksander and Kanazawa, Angjoo and others},
  journal={arXiv preprint arXiv:2310.07204},
  year={2023}
}

@article{barron2022mipnerf360,
    title={Mip-NeRF 360: Unbounded Anti-Aliased Neural Radiance Fields},
    author={Jonathan T. Barron and Ben Mildenhall and 
            Dor Verbin and Pratul P. Srinivasan and Peter Hedman},
    journal={CVPR},
    year={2022}
}

@article{10.1214/aos/1176346577,
author = {J. A. Hartigan and P. M. Hartigan},
title = {{The Dip Test of Unimodality}},
volume = {13},
journal = {The Annals of Statistics},
number = {1},
publisher = {Institute of Mathematical Statistics},
pages = {70 -- 84},
year = {1985},
doi = {10.1214/aos/1176346577},
URL = {https://doi.org/10.1214/aos/1176346577}
}

@article{mueller2022instant,
    author = {Thomas M\"uller and Alex Evans and Christoph Schied and Alexander Keller},
    title = {Instant Neural Graphics Primitives with a Multiresolution Hash Encoding},
    journal = {ACM Trans. Graph.},
    issue_date = {July 2022},
    volume = {41},
    number = {4},
    month = jul,
    year = {2022},
    pages = {102:1--102:15},
    articleno = {102},
    numpages = {15},
    url = {https://doi.org/10.1145/3528223.3530127},
    doi = {10.1145/3528223.3530127},
    publisher = {ACM},
    address = {New York, NY, USA},
}
